\documentclass{INTERSPEECH2023}
\usepackage{todonotes}

\usepackage[acronym]{glossaries}

\newacronym{ASR}{ASR}{Automatic Speech Recognition}
\newacronym{ATS}{ATS}{Automatic Text Summarization}
\newacronym{ASS}{ASS}{Automatic Speech Summarization}

\newacronym{WER}{WER}{Word Error Rate}

\title{BASS: Block-wise Adaptation for Speech Summarization}

\name{Roshan Sharma$^1$, Kenneth Zheng $^1$,Siddhant Arora$^1$,Shinji Watanabe$^1$, Rita Singh$^1$, Bhiksha Raj$^1$}

\address{
  Electrical and Computer Engineering$^1$, Language Technologies Institute $^2$
  \\ Carnegie Mellon University, USA}
\email{\{roshansh,kzheng2,siddhana,swatanab,bhiksha,rsingh\}@cs.cmu.edu}

\interspeechcameraready

\begin{document}

\maketitle
 
\begin{abstract}
End-to-end speech summarization has been shown to improve performance over cascade baselines. However, such models are difficult to train on very large inputs (dozens of minutes or hours) owing to compute restrictions and are hence trained
with truncated model inputs. Truncation leads to poorer models, and a solution to this problem rests in block-wise modeling, i.e., processing a portion of the input frames at a time. In this paper, we develop a method that allows one to train summarization models on very long sequences in an incremental manner. Speech summarization is realized as a streaming process, where hypothesis
summaries are updated every block based on new acoustic information. We devise and test strategies to pass semantic context across the blocks. Experiments on the How2 dataset demonstrate that the proposed block-wise training method improves by 3 points absolute on ROUGE-L over a truncated input baseline.

\end{abstract}

\section{Introduction}

 With the rising amount of data that people consume in daily life -- videos, music, podcasts, meetings, lectures, and more -- building artificial intelligence that can concisely extract important information  ~\cite{sharma2022xnor,palaskar21_interspeech} has gained importance. Speech Summarization refers to the task of developing intelligent machines that can generate condensed textual representations called "summaries" from long audio inputs. Speech summarization, whether extractive \cite{liu2015combining} or abstractive \cite{kano2020asru,kano2022icassp,shon2022slueted}, requires global acoustic context since knowledge of the entire speech signal is helpful for either extracting relevant key-frames or generating comprehensive abstractive summaries.
 

Recently, end-to-end speech summarization models ~\cite{sharma2022end,matsuura2023} have been shown to outperform competitive cascade models that comprise speech recognition and text summarization modules. Such end-to-end models use very long speech sequences as input, and standard transformer models cannot handle very long inputs owing to the quadratic computational complexity of self-attention. Prior work has proposed restricting the scope of attention using the Longformer ~\cite{sharma2022end,beltagy2020longformer} or linear self-attentions like the XNOR-former ~\cite{sharma2022xnor}. However, even with such optimizations, there remains an upper limit on the number of input frames that a given end-to-end model can consume with available computing infrastructure. For example, with a 6-layer conformer~\cite{conformer} encoder, a 32G V-100 GPU can take sequences of length 25,000; and with an XNOR-encoder, the same GPU can take ~45,000 frames. Any input speech sequence of length greater than this upper limit is truncated to be able to train and infer, and truncating inputs makes summarization less accurate since information is effectively removed from the input. Further, attention-based sequence models do not generalize well to input lengths that are different from those used in training ~\cite{deng22b_interspeech}, which makes adapting to longer input sequences important.



 \begin{figure}
    \centering
    \includegraphics[width=0.45\textwidth]{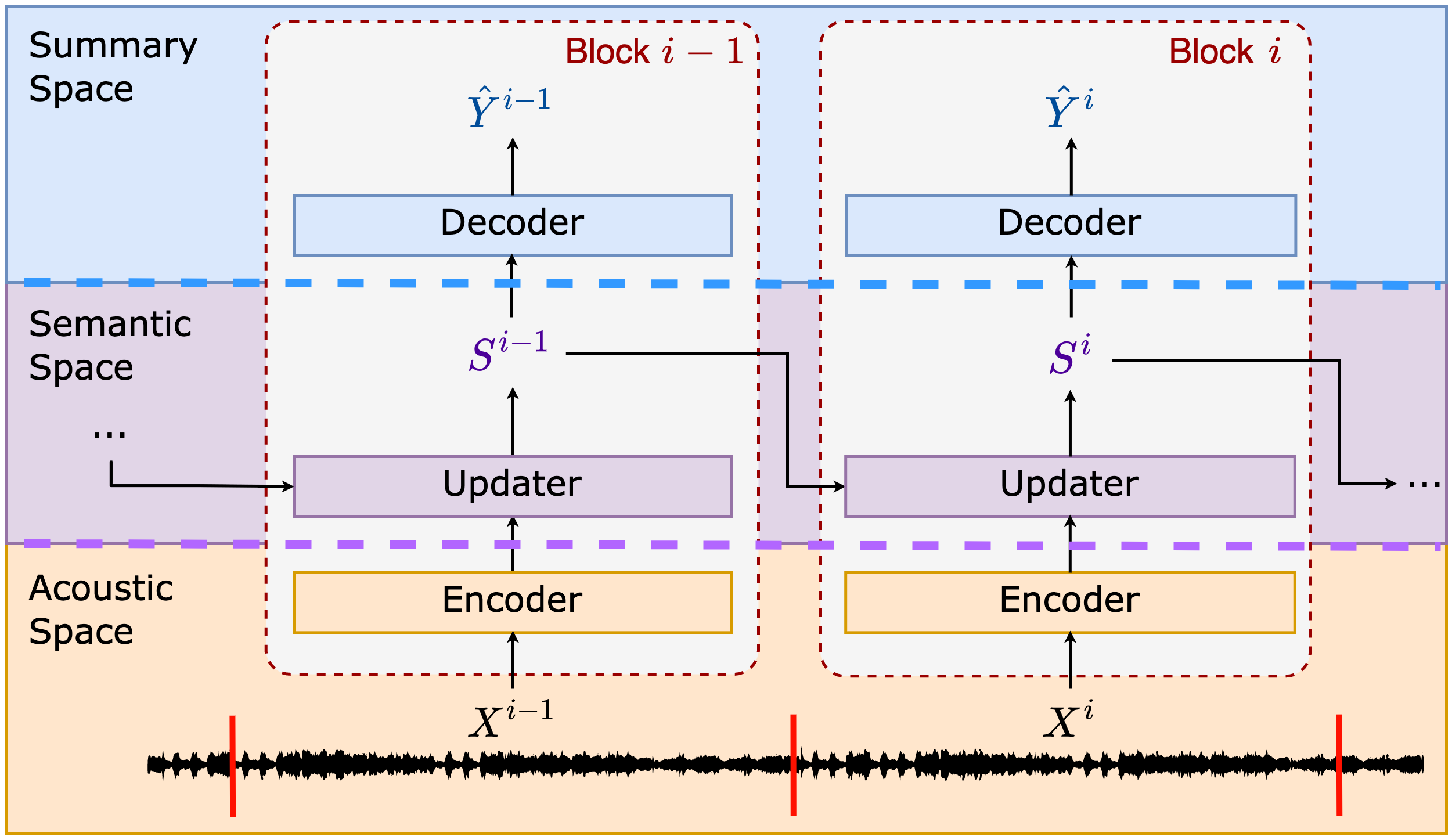}
    \caption{Model architecture of the BASS model. The input audio is split into fixed-size chunks and is processed block-wise. Semantic embeddings are combined between the previous and current blocks using an updater mechanism.}
    \label{fig:block_diagram}
\end{figure}

To address this challenge, one solution is to build models that can process a small set of input frames, i.e., a block of input at a time rather than using the entire input sequence. Such "block-wise" models can be trained in two ways - to predict an output either after seeing multiple blocks of input or after every new block of input. 

Most prior work has been focused on the former to enable streaming inference for tasks like speech recognition ~\cite{rao2017exploring,moritz2020streaming,narayanan2019streaming,tsunoo2021slt,shi2021icassp}, speech translation ~\cite{ma2021translation}, spoken intent detection ~\cite{deng22b_interspeech}, and wake word detection ~\cite{wang2021wake}. Block-wise training for streaming applications uses one block of input at a time to generate a block-level encoding. These block-level encodings from all blocks are then combined to make an utterance-level prediction.  During training, all the inputs from all blocks, intermediate outputs, and the final output are retained in the computational graph for backpropagation. This requires significant compute and memory so these models still can't scale to very long audio sequences.

The latter category of block-wise models overcomes this challenge by producing outputs after every block, so they can be optimized at the block level without requiring the entire input to be present in the computational graph. Prior work in speech recognition for long conversations ~\cite{KimMetze18,hori2021advanced, HoriMHR20} can be considered examples of such block-wise models, since they produce an utterance transcription for every new block of input. Such block-level targets are relatively easy to derive for tasks like speech recognition, where there exists a monotonic alignment between the input frames and output tokens. However, for abstractive speech summarization, the relationship between input frames and output tokens is non-monotonic and indirect, and it is consequently challenging to obtain block-level targets.  
 
 

In this paper, we first mathematically formulate the process of block-wise training and introduce in Figure \ref{fig:block_diagram}  \emph{Block-Wise Adaptation for Speech Summarization (BASS)}, an online model that can be trained with the full reference summary as the block-level target. This means that our model attempts to produce the output summary given only the first block, and then subsequently refines its prediction with every additional block of speech input. While streaming mechanisms assume that new acoustic inputs may incrementally modify the output, we permit the model to modify the entire summary if necessary based on the information present in the new input block. When using such block-wise inputs during training or inference, blocks should have access to the information encoded by previous blocks. 
We propose to achieve this by passing the latent representation across blocks since it is likely where the semantic information is encoded. While it is also possible to carry forward input acoustics or output summaries, these may not be as useful because input acoustics may not be as informative, and output summaries could be erroneous or change entirely with new blocks. In summary, this paper makes the following contributions:


\begin{enumerate}
    \item We introduce Block-wise Adaptation for Speech Summarization (BASS), a novel algorithm for training speech summarization models. BASS predicts a speech summary after consuming a new block of the input speech and allows new summaries to be modified fully if necessary.
    \item We introduce an explicit layer of semantic representation, which aggregates semantics from the input acoustics and is not affected by how it is expressed. We then describe mechanisms to carry this semantic context across blocks for adaptation and training. 
    \item We evaluate the relative strengths of block-wise adaptation from a pretrained model and block-wise training from scratch, and show that BASS improves performance under adaptation settings by 3 points on ROUGE-L. 
\end{enumerate}


\section{Proposed Approach}
\subsection{Formulating Block-wise Training}
\label{section:blockwise_formulation}

Given a long audio instance with N frames of D-dimensional speech features $X=(\mathbf{x}_i \in \mathbb{R}^D | i=1,2,\cdots,N)$, the goal of summarization is to produce a summary token sequence $Y = [y_1,y_2,\cdots y_L]$ of length $L$, which is shorter than the original sequence but still contains the relevant semantic information. 

The input sequence $X$ can also be represented as a sequence of $T$ abutting blocks with block size $B$ such that $X =\{X^1,X^2,...X^T\}$. The $i$-th input block $X^i$ produces a block-level output $\hat{Y}^i$, which is the model hypothesis for the full reference $Y$. We use the notation $X^{1:T}$ to represent $X^1,\cdots,X^T$ and $Y^{1:T}$ to represent $Y^1,\cdots,Y^T$.

The goal of a block-wise model is to generate the best possible summary $\hat{Y}^T$ after seeing the $T$ blocks of input.  Equation \ref{eqn:summing_joint} expresses the probability of observing the final output sequence $Y^T$ given the input blocks $X^{1:T}$ based on the joint conditional density $\mathbb{P}(Y^T,Y^{1:T-1}|X^{1:T})$.
\vspace{0.2pt}
\begin{equation}
   \mathbb{P}(Y^T | X^{1:T}) = \sum_{Y^1}\cdots\sum_{Y^{T-1}} \mathbb{P}(Y^T,Y^{1:T-1}|X^{1:T})
\label{eqn:summing_joint}  
\end{equation}

\noindent Using the chain rule of probability, we can represent the inner term $\mathbb{P}(Y^T,Y^{1:T-1}|X^{1:T})$ as shown in Equation \ref{eqn:expanding_joint}.
\vspace{0.2pt}
\begin{equation}
    \mathbb{P}(Y^{1:T} | X^{1:T})= \mathbb{P}(Y^T|X^{1:T},Y^{1:T-1})\mathbb{P}(Y^{1:T-1}|X^{1:T})
    \label{eqn:expanding_joint}
\end{equation}

\noindent Based on the fact that the model is causal (streaming), the present output cannot depend on future outputs or inputs. This implies $\mathbb{P}(Y^{1:T-1}|X^{1:T}) = \mathbb{P}(Y^{1:T-1}|X^{1:T-1})$. Combining this with Equation \ref{eqn:expanding_joint} results in Equation \ref{eqn:streaming_approx}.


\begin{equation}
\mathbb{P}(Y^{1:T} | X^{1:T}) = \mathbb{P}(Y^T|X^{1:T},Y^{1:T-1})\mathbb{P}(Y^{1:T-1}|X^{1:T-1})
\label{eqn:streaming_approx}
\end{equation}


\noindent This leads to the following general decomposition based on the chain rule and the streaming assumption, shown in Equation \ref{eqn:general_model}. 
\begin{equation}
    P(Y^{1:T} | X^{1:T}) =P(Y^{T}|X^{1:T},Y^{1:T-1}) \cdots P(Y^1|X^1)
\label{eqn:general_model}
\end{equation}
Consider Equation \ref{eqn:summing_joint} which involves marginalizing over the output variables $Y^{1:T-1}$, and is challenging to compute. Rather than evaluating all possible values for past context $Y^{1:T-1}$, we can perform this optimization in a greedy manner, i.e., by choosing the block-level output sequence with the highest probability as context for future predictions. Combining this Viterbi assumption with Equations \ref{eqn:summing_joint} and \ref{eqn:general_model} leads us to the final formulation in Equation \ref{eqn:final_formulation}.
\begin{equation}
    \mathbb{P}(Y^T|X^{1:T}) \approx \max_{Y^T} \mathbb{P}(Y^{T}|X^{1:T},Y^{1:T-1}) \cdots \max_{Y^1} \mathbb{P}(Y^1|X^1) 
\label{eqn:final_formulation}
\end{equation}
In summary, Equation \ref{eqn:final_formulation} demonstrates a setup wherein after receiving a new block of input, we maximize the probability of the block level output being as close as possible to the ground-truth summary given past and current block inputs and past block level outputs. In practice, we take in a block of input and any context from the past, then we compute a divergence between the block-level output and the ground-truth summary. We perform backpropagation with this criterion to update the neural network parameters after every block. 

Apart from the aforementioned assumptions, we can also make the Markov assumption while modeling contextual dependence. To minimize the impact of context from further away blocks on the current block, we can rewrite $\mathbb{P}(Y^{1:i}|X^{1:i}) = \mathbb{P}(Y^{i-M:i}|X^{i-M:i})$.

\subsection{Modeling Strategy and Architecture}
\label{section:general_formulation}

The proposed BASS model is shown in Figure \ref{fig:block_diagram}. Different from past work in summarization, we explicitly introduce a semantic representation variable $S=(\mathbf{s}_i \in \mathbb{R}^F | i=1,2,3,\cdots,M)$, which comprises $M$ $F$-dimensional vectors. $S$ contains the semantic information encoded in the speech $X$, and separates the acoustics from the summary. Modifying the input language or ambient environment changes the acoustics, but not the semantics. Summaries are generated by sampling from this rich semantic representation, and modifying $S$ leads to a different summary $Y$. 


The process of speech summarization occurs at the intersection of three distinctive spaces- the acoustic space, the semantic space, and the summary space. The acoustic space comprises the acoustic input $X$, which it transforms into semantic representations. 
The summary $Y$ can be produced in the summary space by sampling based on the semantic representations $S$. We can reasonably assume that the  acoustics and summary are conditionally independent given the semantics, and thus disentangle the acoustics and the summary.

Consider the task of estimating the most likely summary $Y$ given the input speech $X$ under this setting. 
\vspace{0.1pt}
\begin{equation}
\begin{split}
    \hat{Y} &= \arg\max_Y \mathbb{P}(Y|X) \\
    &= \arg\max_Y \sum_S  \mathbb{P}(Y,S|X) \\  
    &\approx  \arg\max_Y  \max_S P(Y,S|X) 
    \label{eqn:spaces_summarization}
\end{split}
\end{equation}

Equation \ref{eqn:spaces_summarization} describes the process of identifying the most likely hypothesis summary $\hat{Y}$ given the input $X$ and semantic representation $S$.  From Figure \ref{fig:block_diagram}, based on the conditional independence between the summary $Y$ and acoustics $X$ given semantic representation $S$, we can write $\mathbb{P}(Y,S|X)=\mathbb{P}(Y|X)\mathbb{P}(S|X)$. Thus, we can obtain the solution for Equation \ref{eqn:spaces_summarization} using the coordinate descent update shown in Equation \ref{eqn:update_process}.

\begin{equation}
    \begin{split}
        \hat{S} &= \arg\max_S \mathbb{P}(S|X) \\
    \hat{Y} &\approx \arg\max_Y \mathbb{P}(Y|\hat{S}) \max_S \mathbb{P}(S|X) \\ 
    \end{split}
    \label{eqn:update_process}
\end{equation}


From Equations \ref{eqn:update_process} and \ref{eqn:final_formulation}, BASS estimates $\mathbb{P}(Y^i|X^{1:i},Y^{1:i-1})$ using Equation \ref{eqn:blockwise_semantics_eqn}. \begin{equation}
    \mathbb{P}( Y^i | X^{1:i}) = \mathbb{P}(Y^i | S^i) \mathbb{P}(S^i |S^{1:i-1},X^{i})
    \label{eqn:blockwise_semantics_eqn}
\end{equation}

The encoder and decoder model the probabilities $\mathbb{P}(S|X)$ and $\mathbb{P}(Y|S)$ respectively. The updater uses the past semantic embeddings and the current encoder output to produce the current semantic embedding. Figure \ref{fig:updater_mechanisms} shows three alternate structures for our updater to aggregate semantic context from the prior and the current block:


\begin{figure}
    \centering
    \includegraphics[width=0.4\textwidth]{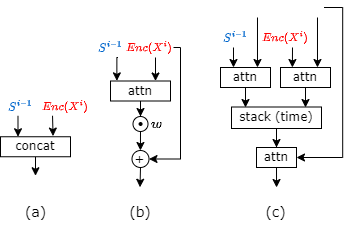}
    \caption{Updater mechanisms (a) concatenation, (b) gated attention, and (c) hierarchical attention use previous embedding $S^{i-1}$ and encoding Enc($X^{i}$) to produce current embedding}
    \label{fig:updater_mechanisms}
\end{figure}

\begin{enumerate}
    \item \textbf{Concatenation}: $S^i = \text{Concat}(S^{i-1}, \text{Encoder}(X^{i}))$ \\
    The current semantic embedding is obtained by concatenating the embeddings from the previous and current blocks.
    \item \textbf{Gated Attention}: \\
    $S^i = \text{Encoder}(X^{i}) + w \cdot \text{Attn}(S^{i-1}, \text{Encoder}(X^{i}))$ \\
    The current and previous semantic embeddings are combined using an attention mechanism and incorporated into the final embedding as a weighted sum.

    \item \textbf{Hierarchical Attention}: \\ $S_i = \text{Attn}([\text{Attn}(S^{i-1}, D^{i}), \text{Attn}(\text{Encoder}(X^{i})), D^{i}]), D^{i})$\\
    This method performs the context passing within each decoder block, based on hierarchical attention \cite{libovicky-helcl-2017-attention}. We compute attention for the current decoder state $D^{i}$ with the previous and current semantic embeddings independently. Then, we stack the two attention outputs and perform a second level of attention between this result and the decoder state.
\end{enumerate}

 \section{Experimental Setup}
 \subsection{Dataset}

\begin{table}[hp]
\centering
\caption{Statistics of the How-2 2,000h Dataset used for model training and evaluation. The maximum input length $N$ (in frames), and maximum output length $L$ (in tokens) are shown.}
\resizebox{0.3\textwidth}{!}{%
\begin{tabular}{lcccc}
\toprule
Set & Max N & Max L & \#Videos \\ \toprule
Train          &      145,082          &    173                                                                &  68,336                              \\ \hline
Test               &     39,537         &  152                                                                  &    2,127                             \\ \hline
\end{tabular}
}
\label{tab:how2_datastats}
\end{table}

The How-2 Dataset \cite{sanabria2018how2} contains 2,000h of instructional videos with corresponding text transcripts, video, speech, translations, and summaries. Abstractive summaries are generated based on user-provided descriptions of the videos. Table \ref{tab:how2_datastats} highlights the number of videos in the train and test partitions of the How2 data. The model features and reference summaries have been made public \footnote{https://github.com/srvk/how2-dataset} by the authors of ~\cite{sharma2022end}.

\subsection{Model Hyperparameters and Evaluation}

\noindent \textbf{Models}: Our models use ESPNet2\footnote{Code will be released in https://github.com/espnet/espnet} ~\cite{watanabe2018espnet}  and are first pre-trained on the ASR task and then fine-tuned for summarization. The encoder consists of convolutional subsampling by factor 4, followed by 12 conformer \cite{conformer} blocks with 8 attention heads and a hidden size of 2048. The decoder has 6 transformer blocks, with 4 attention heads and a hidden size of 2048. The total number of model parameters is 103M. Both the encoder and decoder use a dropout rate of 0.2. We use 43-dimensional filter bank and pitch features as input to the encoder.

\noindent \textbf{ASR}: ASR models are trained with Connectionist Temporal Classification (CTC) and Cross-Entropy loss with CTC weight of 0.3. We use the Adam optimizer with peak lr=0.001, and a warmup scheduler for ASR pre-training. This takes 2 days on 8 V-100 32G GPUs

\noindent \textbf{SUMM}: Our summarization models are trained with cross-entropy loss and label smoothing of 0.15. During inference, we use a beam size of 8. Model averaging was not performed as it was found to hurt summarization performance. Fine-tuning is run for a day on one A40 GPU.

\noindent \textbf{BASS}: For BASS models, we use a block size of 1,000 input frames, corresponding to 10s of audio. We only use the semantic embedding from the previous block as context for the current block unless otherwise specified.

\noindent \textbf{Evaluation}: We evaluate our models with ROUGE \cite{lin-2004-rouge}, METEOR \cite{banerjee-lavie-2005-meteor}, and BERTScore \cite{zhang2020BERTScore}, which are the most common automatic metrics for evaluating summarization models.


\begin{table*}[!thp]
\centering
\caption{Performance of Block-wise Adaptation and Training Approaches compared to Truncated Baselines with different inference strategies using ROUGE, METEOR, and BERTScore metrics- higher scores indicate better performance}
\label{tab:bass_result_summary}
\resizebox{\textwidth}{!}{%
\begin{tabular}{lllllrrrrr}
\toprule
Training Method & Inference & Pre-training & Train Maxlen & Inf. Maxlen & ROUGE-1$\uparrow$ & ROUGE-2$\uparrow$ & ROUGE-L$\uparrow$ & METEOR$\uparrow$ & BERTScore$\uparrow$ \\ \toprule 
Trunc, Restricted Self-Attention~\cite{sharma2022end} & Standard & X & 100s & 100s & 60.73 & 44.9 & 56.10 & 29.30 & 91.53 \\ \hline
Trunc, Full Self-Attention ~\cite{matsuura2023} & Standard & X & 100s & 100s & 65.30 & 51.40 & 62.10 & 32.50 & 93.00 \\
+ TTS Augmentation ~\cite{matsuura2023} & Standard & X & 100s & 100s & 68.40 & 54.10 & 65.00 & 34.90 & 93.80
\\ \hline

Trunc-Baseline  & Standard  & X & 10s & 10s  & 60.87 & 45.12 & 56.79
& 30.00 & 91.6 \\
Trunc-Baseline   & Standard  & X & 30s & 30s & 63.30 & 47.58 & 59.16 & 31.76 & 92.08\\ 
Trunc-Baseline   & Standard  & X & 60s & 60s & 64.57 & 49.11 & 60.49 & 32.47 & 92.38 \\  \hline  


\hline 
BASS-Adapt & Block  & 10s & 30s & 30s  & \textbf{63.99} & \textbf{49.00} & \textbf{60.17} & \textbf{32.17} & \textbf{92.51} \\
BASS-Train & Block & X & 30s & 30s      &       60.87 & 43.12 & 54.79
& 29.12 & 89.40        \\ \bottomrule


\end{tabular}
}
\end{table*}

\begin{table}[ht]
    \centering
    \caption{Part-of-speech coverage between the Predicted Summary and the Reference for Truncated 10s baseline and BASS-ADAPT 30s model}
    \begin{tabular}{lrrrrr}
    \toprule
         Model & Noun & Verb & Adj & Adv & Prop.Noun \\
         \toprule
         Baseline 10-sec & 0.85 & 0.76 & 0.84 & 0.65 & 0.78 \\
        BASS-ADAPT & \textbf{0.87} & \textbf{0.79} & 0.84 & \textbf{0.67} & \textbf{0.81} \\
        \bottomrule
    \end{tabular}
    \label{tab:pos_analysis}
\end{table}

\begin{table}[!ht]
\caption{Performance of BASS models with block level inference across different implementations of the semantic updater model. Models are pre-trained on 10s and fine-tuned on 30s. R-1, R-2, R-3 represent the ROUGE-1, ROUGE-2, and ROUGE-L metrics respectively}
\label{tab:updater_comparison}
\resizebox{\columnwidth}{!}{%
\begin{tabular}{lrrrrr}
    \toprule

Updater    & R-1$\uparrow$   & R-2$\uparrow$   & R-L$\uparrow$   & METEOR$\uparrow$ & BERTScore$\uparrow$ \\ 
\toprule
Concat     & \textbf{63.99} & \textbf{49.00} & \textbf{60.17} & \textbf{32.17}  & \textbf{92.51}    \\ 
Gated Attn & 63.94 &  48.91 &  60.16 &  32.12 & 92.12      \\ 
Hier. Attn & 59.71 & 43.99 &55.74 &  29.32 & 91.27 \\ 
\bottomrule
\end{tabular}
}
\end{table}

\section{Experimental Results}




\subsection{Truncated Input Baselines}
First, we train end-to-end summarization baseline models on truncated inputs (\texttt{Trunc}) that are 10 seconds long and 30 seconds long. Table \ref{tab:bass_result_summary} reports the results of training on truncated inputs and evaluating recordings that are 10 seconds and 30 seconds long, compared to different state-of-the-art approaches referenced in prior work. We note that using the standard full multi-head attention provides significant gains over restricted self-attention, and therefore use the standard multi-head self-attention for our experiments. 


\subsection{Block-wise Training versus Truncated Training}
The proposed BASS method can be used to help models trained on shorter recordings adapt to longer inputs (\texttt{BASS-Adapt}), or to train models from scratch in a block-wise manner (\texttt{BASS-Train}).
Inference for blockwise models can be performed in the \texttt{standard} manner, i.e., where the entire input is fed in at once to predict the final output. Alternatively, \texttt{Block} inference can be performed, where the input is fed as abutting blocks of input as described in Section \ref{section:blockwise_formulation}.   

We train \texttt{BASS-Adapt} initialized from the 10-second truncated baseline to handle 30-second recordings and infer using \texttt{standard} and \texttt{block} mechanisms. \texttt{BASS-Adapt} is compared against \texttt{BASS-Train} by training a model on 30-second recordings from scratch using our BASS algorithm. The latter performs worse - for training from scratch, the challenge is relatively poor initial context. Initially, the learned context is not very helpful which leads to slower convergence and poorer performance. 

We compare both model adaptation and training strategies in Table \ref{tab:bass_result_summary}. We see that the proposed \texttt{BASS-Adapt} approach outperforms \texttt{BASS-Train} on all metrics. We also observe that our proposed BASS algorithm improves over the truncated 10-second baseline and the truncated 30-second baseline. \texttt{BASS-Adapt} with block inference results in a nearly 4-point improvement in ROUGE-L over the 10-second truncated baseline, and a 1-point improvement in ROUGE-L over the truncated 30-second baseline. This result is comparable to that obtained by a truncated input baseline that takes in 60 seconds of audio, showing that our BASS model trained with 30-second recordings comprising 10-second chunks can do as well as a model trained on 60-second recordings.  The proposed approach is more computationally efficient than the baseline by a factor of 3 since the proposed approach uses 3x smaller inputs 3 times for quadratic self-attentions.


Finally, Table \ref{tab:pos_analysis} sheds light on the improvement in the prediction of different parts of speech in the reference summary using the best \texttt{BASS-ADAPT} model. We observe that the proposed model generally improves the prediction of all parts of speech. Future work may benefit from exploring named entity prediction for summaries.


\subsection{Comparison of Block-wise Adaptation Strategies}

Table \ref{tab:updater_comparison} compares the various modeling strategies for the semantic updater. We observe that simply concatenating the semantic embedding of the previous (one) block with the current block yields significant improvements in summarization performance. Of the three updater mechanisms described in Figure \ref{fig:updater_mechanisms}, gated attention and concatenation appear to yield similar gains in performance, with hierarchical attention performing significantly worse. Gated attention is able to achieve similar performance while reducing having a very small memory footprint compared to concatenation. \\



\section{Conclusion}
In this paper, we address the challenge of training end-to-end speech summarization models over very long inputs. Though certain optimizations can be used to improve the upper limit on input size for summarization models, performance is limited by the truncation of model inputs during training and inference. We propose Block-wise Adaptive for Speech Sequences (BASS) to address this challenge - an algorithm that consumes the input in blocks and passes semantic context across blocks to encourage better learning.  The BASS algorithm can be used to adapt pre-trained truncated input models to longer sequences, or train models over long sequences from scratch. We show that the proposed model outperforms truncated baselines and enables the training of speech summarization models with very long inputs.


\section{Acknowledgements}
We would like to thank Raphael Olivier, Hira Dhamyal and Mark Lindsey for their helpful feedback. This work used PSC Bridges2 and NCSA Delta through allocations CIS210014 and IRI120015 from the Advanced Cyberinfrastructure Coordination Ecosystem: Services \& Support (ACCESS) program,
which is supported by National Science Foundation grants
\#2138259, \#2138286, \#2138307, \#2137603, and \#2138296.

\newpage

\bibliographystyle{IEEEtran}
\bibliography{main}

\begin{thebibliography}{10}
\providecommand{\url}[1]{#1}
\csname url@samestyle\endcsname
\providecommand{\newblock}{\relax}
\providecommand{\bibinfo}[2]{#2}
\providecommand{\BIBentrySTDinterwordspacing}{\spaceskip=0pt\relax}
\providecommand{\BIBentryALTinterwordstretchfactor}{4}
\providecommand{\BIBentryALTinterwordspacing}{\spaceskip=\fontdimen2\font plus
\BIBentryALTinterwordstretchfactor\fontdimen3\font minus
  \fontdimen4\font\relax}
\providecommand{\BIBforeignlanguage}[2]{{%
\expandafter\ifx\csname l@#1\endcsname\relax
\typeout{** WARNING: IEEEtran.bst: No hyphenation pattern has been}%
\typeout{** loaded for the language `#1'. Using the pattern for}%
\typeout{** the default language instead.}%
\else
\language=\csname l@#1\endcsname
\fi
#2}}
\providecommand{\BIBdecl}{\relax}
\BIBdecl

\bibitem{sharma2022xnor}
R.~Sharma and B.~Raj, ``Xnor-former: Learning accurate approximations in long
  speech transformers,'' \emph{arXiv preprint arXiv:2210.16643}, 2022.

\bibitem{palaskar21_interspeech}
S.~Palaskar, R.~Salakhutdinov, A.~W. Black, and F.~Metze, ``{Multimodal Speech
  Summarization Through Semantic Concept Learning},'' in \emph{Proc.
  Interspeech 2021}, 2021, pp. 791--795.

\bibitem{liu2015combining}
S.-H. Liu, K.-Y. Chen, B.~Chen, H.-M. Wang, H.-C. Yen, and W.-L. Hsu,
  ``Combining relevance language modeling and clarity measure for extractive
  speech summarization,'' \emph{IEEE/ACM Transactions on Audio, Speech, and
  Language Processing}, vol.~23, no.~6, pp. 957--969, 2015.

\bibitem{kano2020asru}
T.~Kano, A.~Ogawa, M.~Delcroix, and S.~Watanabe, ``Attention-based
  multi-hypothesis fusion for speech summarization,'' in \emph{2021 IEEE
  Automatic Speech Recognition and Understanding Workshop (ASRU)}, 2021, pp.
  487--494.

\bibitem{kano2022icassp}
------, ``Integrating multiple asr systems into nlp backend with attention
  fusion,'' in \emph{ICASSP 2022 - 2022 IEEE International Conference on
  Acoustics, Speech and Signal Processing (ICASSP)}, 2022, pp. 6237--6241.

\bibitem{shon2022slueted}
\BIBentryALTinterwordspacing
S.~Shon, S.~Arora, C.-J. Lin, A.~Pasad, F.~Wu, R.~Sharma, W.-L. Wu, H.-Y. Lee,
  K.~Livescu, and S.~Watanabe, ``Slue phase-2: A benchmark suite of diverse
  spoken language understanding tasks,'' 2022. [Online]. Available:
  \url{https://arxiv.org/abs/2212.10525}
\BIBentrySTDinterwordspacing

\bibitem{sharma2022end}
R.~Sharma, S.~Palaskar, A.~W. Black, and F.~Metze, ``End-to-end speech
  summarization using restricted self-attention,'' in \emph{ICASSP 2022-2022
  IEEE International Conference on Acoustics, Speech and Signal Processing
  (ICASSP)}.\hskip 1em plus 0.5em minus 0.4em\relax IEEE, 2022, pp. 8072--8076.

\bibitem{matsuura2023}
\BIBentryALTinterwordspacing
K.~Matsuura, T.~Ashihara, T.~Moriya, T.~Tanaka, A.~Ogawa, M.~Delcroix, and
  R.~Masumura, ``Leveraging large text corpora for end-to-end speech
  summarization,'' 2023. [Online]. Available:
  \url{https://arxiv.org/abs/2303.00978}
\BIBentrySTDinterwordspacing

\bibitem{beltagy2020longformer}
I.~Beltagy, M.~E. Peters, and A.~Cohan, ``Longformer: The long-document
  transformer,'' \emph{arXiv preprint arXiv:2004.05150}, 2020.

\bibitem{conformer}
A.~Gulati, J.~Qin, C.-C. Chiu, N.~Parmar, Y.~Zhang, J.~Yu, W.~Han, S.~Wang,
  Z.~Zhang, Y.~Wu, and R.~Pang, ``{Conformer: Convolution-augmented Transformer
  for Speech Recognition},'' in \emph{Proc. Interspeech 2020}, 2020, pp.
  5036--5040.

\bibitem{deng22b_interspeech}
K.~Deng, S.~Watanabe, J.~Shi, and S.~Arora, ``{Blockwise Streaming Transformer
  for Spoken Language Understanding and Simultaneous Speech Translation},'' in
  \emph{Proc. Interspeech 2022}, 2022, pp. 1746--1750.

\bibitem{rao2017exploring}
K.~Rao, H.~Sak, and R.~Prabhavalkar, ``Exploring architectures, data and units
  for streaming end-to-end speech recognition with rnn-transducer,'' in
  \emph{2017 IEEE Automatic Speech Recognition and Understanding Workshop
  (ASRU)}.\hskip 1em plus 0.5em minus 0.4em\relax IEEE, 2017, pp. 193--199.

\bibitem{moritz2020streaming}
N.~Moritz, T.~Hori, and J.~Le, ``Streaming automatic speech recognition with
  the transformer model,'' in \emph{ICASSP 2020-2020 IEEE International
  Conference on Acoustics, Speech and Signal Processing (ICASSP)}.\hskip 1em
  plus 0.5em minus 0.4em\relax IEEE, 2020, pp. 6074--6078.

\bibitem{narayanan2019streaming}
A.~Narayanan, R.~Prabhavalkar, C.-C. Chiu, D.~Rybach, T.~N. Sainath, and
  T.~Strohman, ``Recognizing long-form speech using streaming end-to-end
  models,'' in \emph{2019 IEEE Automatic Speech Recognition and Understanding
  Workshop (ASRU)}, 2019, pp. 920--927.

\bibitem{tsunoo2021slt}
E.~Tsunoo, Y.~Kashiwagi, and S.~Watanabe, ``Streaming transformer asr with
  blockwise synchronous beam search,'' in \emph{2021 IEEE Spoken Language
  Technology Workshop (SLT)}, 2021, pp. 22--29.

\bibitem{shi2021icassp}
Y.~Shi, Y.~Wang, C.~Wu, C.-F. Yeh, J.~Chan, F.~Zhang, D.~Le, and M.~Seltzer,
  ``Emformer: Efficient memory transformer based acoustic model for low latency
  streaming speech recognition,'' in \emph{ICASSP 2021 - 2021 IEEE
  International Conference on Acoustics, Speech and Signal Processing
  (ICASSP)}, 2021, pp. 6783--6787.

\bibitem{ma2021translation}
X.~Ma, Y.~Wang, M.~J. Dousti, P.~Koehn, and J.~Pino, ``Streaming simultaneous
  speech translation with augmented memory transformer,'' in \emph{ICASSP 2021
  - 2021 IEEE International Conference on Acoustics, Speech and Signal
  Processing (ICASSP)}, 2021, pp. 7523--7527.

\bibitem{wang2021wake}
Y.~Wang, H.~Lv, D.~Povey, L.~Xie, and S.~Khudanpur, ``Wake word detection with
  streaming transformers,'' in \emph{ICASSP 2021-2021 IEEE International
  Conference on Acoustics, Speech and Signal Processing (ICASSP)}.\hskip 1em
  plus 0.5em minus 0.4em\relax IEEE, 2021, pp. 5864--5868.

\bibitem{KimMetze18}
\BIBentryALTinterwordspacing
S.~Kim and F.~Metze, ``Dialog-context aware end-to-end speech recognition,'' in
  \emph{2018 {IEEE} Spoken Language Technology Workshop, {SLT} 2018, Athens,
  Greece, December 18-21, 2018}.\hskip 1em plus 0.5em minus 0.4em\relax {IEEE},
  2018, pp. 434--440. [Online]. Available:
  \url{https://doi.org/10.1109/SLT.2018.8639044}
\BIBentrySTDinterwordspacing

\bibitem{hori2021advanced}
T.~Hori, N.~Moritz, C.~Hori, and J.~L. Roux, ``{Advanced Long-Context
  End-to-End Speech Recognition Using Context-Expanded Transformers},'' in
  \emph{Proc. Interspeech 2021}, 2021, pp. 2097--2101.

\bibitem{HoriMHR20}
T.~Hori, N.~Moritz, C.~Hori, and J.~Le~Roux, ``Transformer-based long-context
  end-to-end speech recognition,'' \emph{Proc. Interspeech 2020}, pp.
  5011--5015, 2020.

\bibitem{libovicky-helcl-2017-attention}
\BIBentryALTinterwordspacing
J.~Libovick{\'y} and J.~Helcl, ``Attention strategies for multi-source
  sequence-to-sequence learning,'' in \emph{Proceedings of the 55th Annual
  Meeting of the Association for Computational Linguistics (Volume 2: Short
  Papers)}.\hskip 1em plus 0.5em minus 0.4em\relax Vancouver, Canada:
  Association for Computational Linguistics, Jul. 2017, pp. 196--202. [Online].
  Available: \url{https://aclanthology.org/P17-2031}
\BIBentrySTDinterwordspacing

\bibitem{sanabria2018how2}
\BIBentryALTinterwordspacing
R.~Sanabria, O.~Caglayan, S.~Palaskar, D.~Elliott, L.~Barrault, L.~Specia, and
  F.~Metze, ``How2: a large-scale dataset for multimodal language
  understanding,'' \emph{arXiv preprint arXiv:1811.00347}, 2018. [Online].
  Available: \url{https://arxiv.org/abs/1811.00347}
\BIBentrySTDinterwordspacing

\bibitem{watanabe2018espnet}
\BIBentryALTinterwordspacing
S.~Watanabe, T.~Hori, S.~Karita, T.~Hayashi, J.~Nishitoba, Y.~Unno, N.~{Enrique
  Yalta Soplin}, J.~Heymann, M.~Wiesner, N.~Chen, A.~Renduchintala, and
  T.~Ochiai, ``{ESPnet}: End-to-end speech processing toolkit,'' in
  \emph{Proceedings of Interspeech}, 2018, pp. 2207--2211. [Online]. Available:
  \url{http://dx.doi.org/10.21437/Interspeech.2018-1456}
\BIBentrySTDinterwordspacing

\bibitem{lin-2004-rouge}
\BIBentryALTinterwordspacing
C.-Y. Lin, ``{ROUGE}: A package for automatic evaluation of summaries,'' in
  \emph{Text Summarization Branches Out}.\hskip 1em plus 0.5em minus
  0.4em\relax Barcelona, Spain: Association for Computational Linguistics, Jul.
  2004, pp. 74--81. [Online]. Available:
  \url{https://aclanthology.org/W04-1013}
\BIBentrySTDinterwordspacing

\bibitem{banerjee-lavie-2005-meteor}
\BIBentryALTinterwordspacing
S.~Banerjee and A.~Lavie, ``{METEOR}: An automatic metric for {MT} evaluation
  with improved correlation with human judgments,'' in \emph{Proceedings of the
  {ACL} Workshop on Intrinsic and Extrinsic Evaluation Measures for Machine
  Translation and/or Summarization}.\hskip 1em plus 0.5em minus 0.4em\relax Ann
  Arbor, Michigan: Association for Computational Linguistics, Jun. 2005, pp.
  65--72. [Online]. Available: \url{https://aclanthology.org/W05-0909}
\BIBentrySTDinterwordspacing

\bibitem{zhang2020BERTScore}
\BIBentryALTinterwordspacing
T.~Zhang*, V.~Kishore*, F.~Wu*, K.~Q. Weinberger, and Y.~Artzi, ``Bertscore:
  Evaluating text generation with bert,'' in \emph{International Conference on
  Learning Representations}, 2020. [Online]. Available:
  \url{https://openreview.net/forum?id=SkeHuCVFDr}
\BIBentrySTDinterwordspacing

\end{thebibliography}

\end{document}